\documentclass[10pt,twocolumn,letterpaper]{article}
\PassOptionsToPackage{table}{xcolor}
\usepackage{authblk}
\usepackage[pagenumbers]{cvpr}
\usepackage{microtype}
\usepackage{multirow}
\usepackage{makecell}
\usepackage{placeins}
\usepackage{latexml}
\definecolor{cvprblue}{rgb}{0.21,0.49,0.74}
\usepackage[breaklinks,colorlinks,allcolors=cvprblue,hyperfootnotes=false]{hyperref}
\usepackage[capitalize,noabbrev]{cleveref}
\hypersetup{
  pdftitle={VeriVLA: Adaptive Test-Time Verification for Flow-Based VLAs with Latent World Models},
  pdfauthor={Xueli Sun, Yuxiang Lu, Yantong Liu, Xianzhe Fan, Xi Chen, Zhe Liu, Hengshuang Zhao},
  pdfsubject={Adaptive test-time verification for vision-language-action policies},
  pdfkeywords={robotics, vision-language-action models, latent world models, test-time verification}
}
\AtBeginEnvironment{abstract}{\hyphenpenalty=10000\exhyphenpenalty=50\emergencystretch=1em}

\title{VeriVLA: Adaptive Test-Time Verification for Flow-Based VLAs\\with Latent World Models}
\renewcommand{\Authfont}{\large}
\renewcommand{\Affilfont}{\normalsize}
\renewcommand{\Authsep}{\quad}
\renewcommand{\Authand}{\quad}
\renewcommand{\Authands}{\quad}
\author{Xueli Sun}
\author{Yuxiang Lu}
\author{Yantong Liu}
\author{Xianzhe Fan\authorcr}
\author{Xi Chen}
\author{Zhe Liu\textsuperscript{\textdagger}}
\author{Hengshuang Zhao\textsuperscript{*}}
\iflatexml
  \affil{The University of Hong Kong}
\else
  \affil{The University of Hong Kong\\
  {\small \textsuperscript{*}Corresponding author.\quad \textsuperscript{\textdagger}Project lead.}}
\fi
\iflatexml
  \date{\textsuperscript{*}Corresponding author.\quad \textsuperscript{\textdagger}Project lead.}
\else
  \date{}
\fi
\begin{document}
\maketitle

\begin{abstract}

Flow-based Vision-Language-Action (VLA) policies are inherently stochastic: different noise seeds can produce action chunks of varying quality from the same observation. At test time, however, the policy typically commits to a single sample without an explicit estimate of its quality. In fine-grained manipulation, executing a suboptimal action can irreversibly disrupt task progress and lead to failure. To exploit policy stochasticity, we introduce a training-free two-axis trigger that uses two initial samples to measure within-chunk temporal variation and cross-sample variance, detecting the critical states. Once triggered, a JEPA-based latent world model ranks candidate action chunks by predicting action-conditioned future latent states and the corresponding changes in task progress. By operating in latent space, the verifier avoids pixel reconstruction and enables efficient and more accurate ranking. Across three real-world tasks~(\textit{i.e.}, pen insertion, pipetting, and apple placement), our method raises the average completion score from 50.8\% to 65.5\%, while adding 326 ms for sampling and verification per triggered decision on a single RTX 4090. Consistent gains are also observed in simulation, where the average success rate rises from 66.4\% to 73.6\% in the clean setting of RoboTwin 2.0, from 62.8\% to 69.6\% in its randomized setting, and from 96.9\% to 97.6\% on LIBERO. Project page: \url{https://shirley430316.github.io/projects/verivla/}.

\end{abstract}

\section{Introduction}
\label{sec:intro}

Vision-Language-Action (VLA) models have emerged as a promising paradigm towards general robotic control \cite{brohan2023rt1roboticstransformerrealworld,kim2024openvlaopensourcevisionlanguageactionmodel, black2024pi0visionlanguageactionflowmodel}. In the pretraining stage, the field has evolved from VLMs autoregressively generating over discretized action tokens \cite{brohan2023rt2visionlanguageactionmodelstransfer,
kim2024openvlaopensourcevisionlanguageactionmodel, pertsch2025fastefficientactiontokenization} to more recent flow-matching action experts coupled with pretrained VLMs \cite{
nvidia2025gr00tn1openfoundation,
black2024pi0visionlanguageactionflowmodel, intelligence2025pi05visionlanguageactionmodelopenworld}, enabling scalable learning from diverse action trajectories. The paradigm shift was foreshadowed by diffusion-based visuomotor policies \cite{chi2024diffusionpolicyvisuomotorpolicy}, which established iterative denoising as an effective means of learning multi-modal (\textit{i.e.}, multi-peaked) action distributions. However, the stochasticity that benefits learning also introduces a unique challenge at deployment.

\begin{figure}[t!]

    \centering

    \includegraphics[width=1\linewidth]{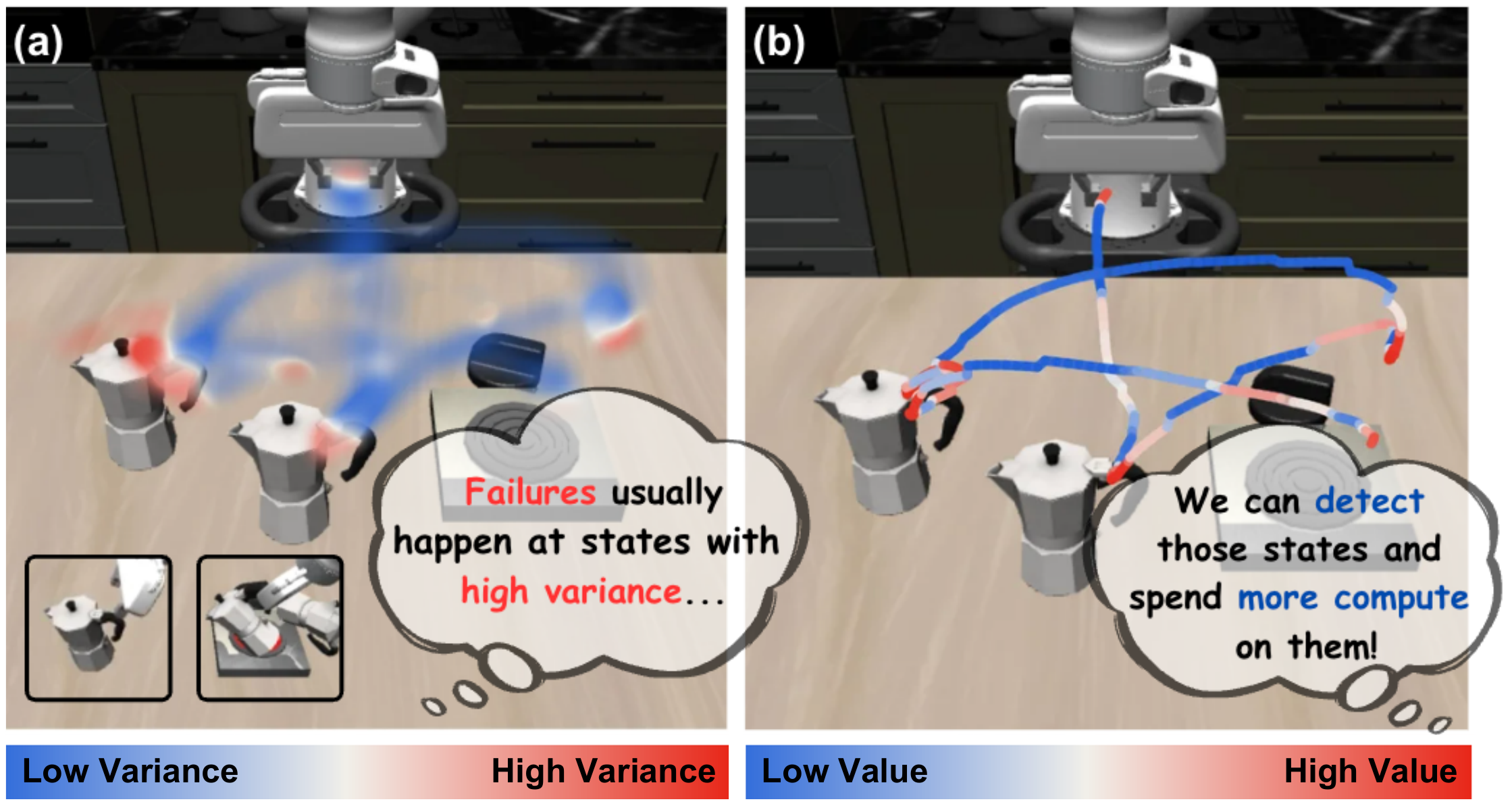}

    \caption{\textbf{(a) Policy variance in the LIBERO environment.} Repeated sampling reveals higher action variance in contact-rich regions, where task failures are most likely to occur. \textbf{(b) Values of the proposed trigger.} The trigger value aligns with task difficulty and the policy's internal uncertainty. These observations motivate triggering repeated sampling and verification at critical decision steps.}

    \label{fig:intuition}

\end{figure}

\begin{figure*}[t]

    \centering

    \includegraphics[width=0.95\linewidth]{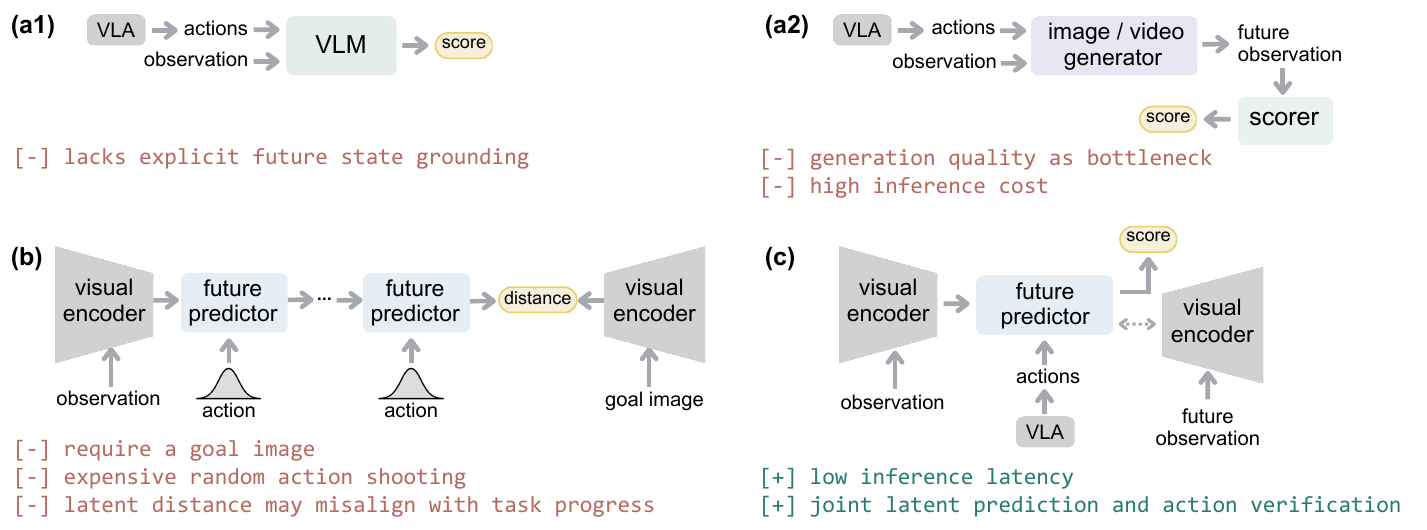}

    \caption{\textbf{Conceptual comparison between different paradigms.} \textbf{(a1) Sampling-and-selection with VLM:}  Directly predict the delta progress with the current observation and the action chunk produced by VLAs.
    \textbf{(a2) Sampling-and-selection with generative models:}  Extract the delta progress after generating the estimated future. \textbf{(b) Model Predictive Control (MPC) with latent world models:} Execute action(s) that minimize the latent distance with the goal image.
    \textbf{(c) Latent world model verifier (ours):} Estimate the future latent and the delta progress in a unified latent space.}

    \label{fig:concept_comp}

\end{figure*}
Given the same observation and instruction, different noise seeds yield samples of different quality during inference. However, the flow-based VLA model cannot distinguish which action chunk is most likely to succeed. During execution, a robot must commit to one sampled action chunk, which may irreversibly change the physical world. A good action chunk advances the task, while a poor one could directly lead to failure or accumulate errors that drift the trajectory out of distribution \cite{ross2011reductionimitationlearningstructured, liu2026streampi, lu2026faster}. We empirically show that the policy variance is non-uniform along the task, and failures tend to happen in high-variance segments of the action trajectory, as in Figure~\ref{fig:intuition}(a). What if we could detect these critical moments before committing to an irreversible action, and briefly ``rewind'' the decision by spending additional computation? This calls for two components: a trigger that decides \textit{when} the states are critical and additional computation is worthwhile, and a verifier that determines \textit{which} sampled action should be executed.

To answer the first question, we start with the policy's uncertainty. As shown in \Cref{fig:intuition}(a), when the end-effector is close to a contact-rich interaction region, the trajectory follows a more complex motion \cite{pang2023globalplanningcontactrichmanipulation}, and internally, the policy uncertainty increases. We therefore combine temporal action variation and cross-sample uncertainty into a training-free two-axis trigger. Computed with two sampled action chunks, the trigger closely tracks the contact information from the environment and aligns well with the policy's intrinsic uncertainty, as shown in \Cref{fig:intuition}(b). We therefore use the trigger to activate repeated sampling and verification at critical decision steps.

Once repeated sampling is activated, the second question arises: how should candidate actions be evaluated? We argue that latent world models are a natural fit. The verifier evaluates the quality of action chunk samples by predicting the progress or relative ranking \cite{kwok2025robomonkeyscalingtesttimesampling, fan2026dreamavoidcriticalphasetesttimedreaming}. Therefore, the model must understand world dynamics and represent the future states resulting from executing the action chunk.
In addition, verification is a low-dimensional problem: how much an action chunk advances the task is a scalar judgment, and most camera pixels are irrelevant to it. As a result, pixel-level world models have two pitfalls: (1) the image or video generation incurs substantial latency, making it difficult to deploy at real-time control frequencies; and (2) they solve the harder pixel generation before the easier progress estimation, making the quality of downstream verification unnecessarily dependent on generation fidelity. We compare these paradigms in \Cref{fig:concept_comp}. This motivates verification in a latent space, which compresses the pixel observation into compact and dynamic-rich features, where we can jointly predict the latent states and task progress.

Building on these observations, we recast action randomness not as a hazard to be suppressed but as a resource to select from, with the latent world model as a verifier. Specifically, we claim the following contributions:

\begin{enumerate}

\item \textbf{A latent world model as a verifier.} We exploit the action-conditioned latent representation produced by a pretrained latent world model to estimate the progress of candidate action chunks.

\item \textbf{An empirical characterization of policy stochasticity in flow-based VLAs and the two-axis trigger.} We characterize policy stochasticity. Leveraging the temporal action variation and the cross-sampling variance, we propose a two-axis trigger that is action-based and training-free. It detects the critical states and reduces inference costs while preserving the benefits of test-time selection.

\item \textbf{A simple training, evaluation, and real-robot deployment pipeline.} On real-world manipulation tasks, each triggered round of repeated sampling and verification adds 326 ms of latency. Our method improves the average completion score by 14.7 percentage points over the base policy, outperforming baselines, including VLM-based verifiers and pixel-level world model verifiers.

\end{enumerate}
\section{Related Work}

\paragraph{Test-time Enhancement for VLAs.}
Existing approaches can be broadly categorized according to whether the VLA policy is updated at deployment time.
Test-time training methods, such as EVOLVE-VLA \cite{bai2025evolvevlatesttimetrainingenvironment} and TT-VLA \cite{liu2026ontheflyvlaadaptationtesttime}, perform online policy optimization during deployment using dense reward signals. Rather than relying on external reward estimation, T$^2$VLA \cite{chen2026trustinstinctsconfidencedriventesttime} derives intrinsic rewards from the policy's own confidence to enable self-supervised adaptation.

Another line of work keeps the pretrained VLA policy frozen and improves action quality through either gradient-free selection or gradient-based augmentation. Gradient-free methods \cite{xu2026vgasvalueguidedactionchunkselection} generally adopt a repeated-sampling-and-selection strategy where a model-based verifier evaluates candidate actions. RoboMonkey \cite{kwok2025robomonkeyscalingtesttimesampling} uses a VLM-based verifier to select the optimal action, demonstrating a test-time scaling law of autoregressive VLAs.  GPC-RANK \cite{qi2026inferencetimeenhancementgenerativerobot} employs a lightweight pixel-generation model with a downstream estimator to rank candidate actions. DreamAvoid \cite{fan2026dreamavoidcriticalphasetesttimedreaming} combines a pretrained foundation-level, action-conditioned video generation model with an estimator and introduces a vision-based trigger to reduce computational overhead. In contrast, gradient-based methods such as Guided Action Flow \cite{yang2026guidedactionflowqguided} backpropagate gradients from the verifier score to refine sampled actions.

In this work, we follow the repeated-sampling-and-selection paradigm. Unlike prior verifier-based methods, which rely on VLMs or pixel-level world models followed by downstream signal extraction, our approach uses a latent world model to jointly predict dynamics and estimate task-specific progress, synergizing with an action-based trigger to allocate computation adaptively. We compare more test-time verification methods in Appendix~\ref{appendix:related_work}.

\vspace{-3pt}
\paragraph{Latent World Models for Robotics.}
Building on advances in visual representation learning, prior research on latent world models has developed along two main directions. The first line of work \cite{sobal2025learningrewardfreeofflinedata,maes2026leworldmodelstableendtoendjointembedding,zhang2026hierarchicalplanninglatentworld}, represented by DINO-WM \cite{zhou2025dinowmworldmodelspretrained}, V-JEPA 2-AC \cite{assran2025vjepa2selfsupervisedvideo} and H-JEPA \cite{zhang2026hjepaendtoendlearninghierarchical} utilizes pretrained visual encoders and trains dynamics predictors to perform zero-shot planning in latent space. Given a goal image, these methods employ search-based algorithms to identify suitable action proposals. The second direction combines latent world models with VLAs. VLA-JEPA \cite{sun2026vlajepaenhancingvisionlanguageactionmodel}, Being-H0.7 \cite{luo2026beingh07latentworldactionmodel} and Copper-Policy \cite{feng2026copperpolicyfocusrepresentationrobust} incorporate the prediction objective of latent world models into VLA training, while LaWAM \cite{chen2026lawamlatentworldaction} introduces a latent subgoal as an additional conditioning signal. These integrated approaches improve overall manipulation performance and generalization.
\section{Method}

\begin{figure*}[t]
    \centering
    \includegraphics[width=1\linewidth]{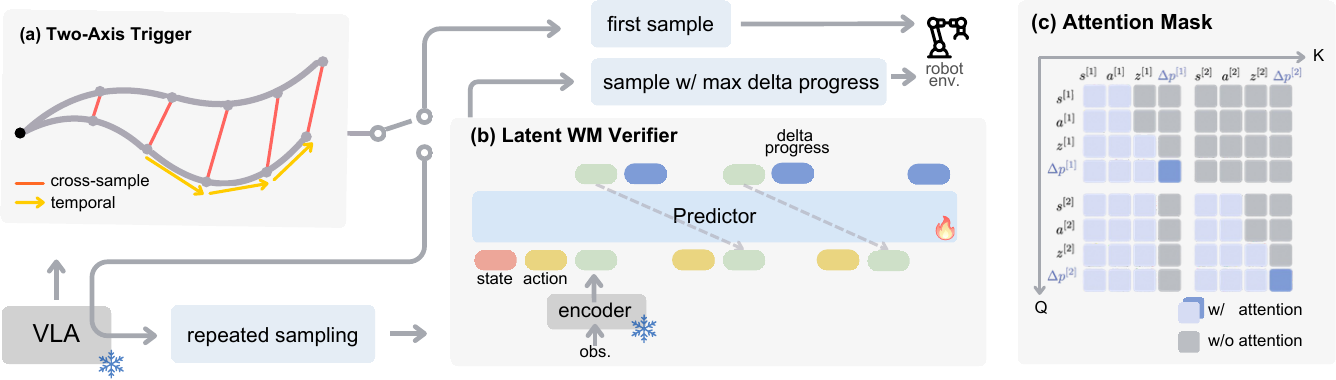}

    \caption{\textbf{Overall pipeline.} \textbf{(a) Two-Axis Trigger.} The VLA policy generates two initial action chunk samples, based on which the two-axis trigger decides whether to activate repeated sampling. \textbf{(b) Latent World Model Verifier.} If triggered, the verifier evaluates $K$ candidate chunks and selects the one with the highest estimated delta progress for execution. The JEPA-based latent world model autoregressively takes states, actions, and observations as input and predicts delta progress in the latent space. \textbf{(c) Attention Mask.} The delta progress key is designed to be invisible from other queries to remove possible task-specific shortcuts for world modeling.}
    \label{fig:method}

\end{figure*}

\subsection{Problem Formulation}

In the standard VLA setting, at each decision step $t$, a policy $\pi_\theta$ generates an action chunk with current observation $o_t$, language instruction $l$, and robot proprioceptive state $s_t$:
\begin{equation}
    a_{t:t+H} \sim \pi_\theta(\cdot \mid o_t, l, s_t),
\end{equation}
where $a_{t:t+H}$ denotes $H$ consecutive actions, from timestep $t$ to $t+H-1$. Recent VLAs commonly employ flow-matching-based action generation \cite{intelligence2025pi05visionlanguageactionmodelopenworld}, which takes random Gaussian noise as input. This noise dependence is typically omitted from the policy notation. We make this dependence explicit:
\begin{equation}
    a^{(i)}_{t:t+H} \sim \pi_\theta(\cdot \mid o_t, l, s_t, N_i),
\end{equation}
where $N_i$ denotes the random seed used to generate the Gaussian noise, with a distinct seed for each sampling index $i$. Repeated sampling therefore produces a set of diverse action chunks $\{a_{t:t+H}^{(i)}\}_{i=1}^K$.

Executing an action chunk changes task progress; therefore we define the quality of an action chunk as its change in progress, referred to as \textit{delta progress},
\begin{equation}
    \Delta p = p_{t+H} - p_t,
\end{equation}
where $p_t$ is the task progress at timestep $t$. Compared with absolute progress, delta progress matches the effect of the action chunk, making it a more direct and intuitive target for action evaluation \cite{chen2025robotwin20scalabledata}.

\subsection{Overview}

As illustrated in \Cref{fig:method}, our framework comprises two tightly related components: Two-Axis Trigger and Latent World Model Verifier. Based on initial action chunk samples, the trigger identifies decision steps at which the policy is likely to produce unreliable actions and activates verification only in these challenging situations.

When the trigger is inactive, no additional action chunks are sampled, and the initial sample $a^{(1)}_{t:t+H}$ is executed directly. When the trigger is activated, the policy generates additional samples to expand the candidate action pool. The verifier $V_\phi$ then predicts the delta progress associated with each candidate:
\begin{equation}
    \left(\hat{z}_{t+H}^{(i)}, \Delta \hat{p}^{(i)}\right)
    = V_\phi\left(s_t, a^{(i)}_{t:t+H}, o_t; q_l\right),
\end{equation}
where $q_l$ is the learned query token for the instructed task, $\hat{z}_{t+H}^{(i)}$ denotes the predicted visual latent at time step $t+H$, and $\Delta \hat{p}^{(i)}$ denotes the estimated delta progress of the sample $i$. The action chunk with the largest estimated delta progress is subsequently executed.

\subsection{Two-axis Trigger}
\label{sec:trigger}

Following prior work \cite{hu2024adaflowimitationlearningvarianceadaptive, fan2026dreamavoidcriticalphasetesttimedreaming} and our empirical observations (shown in \Cref{fig:intuition}), we design two complementary signals for allocating verification computation: (C1) activate in contact-rich regions, where manipulation is inherently more complex \cite{pang2023globalplanningcontactrichmanipulation} and failures concentrate; and (C2) activate at states where repeated sampling produces diverse candidate actions, providing the verifier with meaningful alternatives to choose from.

At timestep $t$, we draw $k$ action chunks $\{a_{t:t+H}^{(i)}\}_{i=1}^{k}$ from the VLA policy, where $k$ is chosen to be much smaller than the full sampling budget $K$ used during verification, so that the trigger incurs acceptable overhead. To avoid bias toward fast-moving stages, each action chunk is first normalized by its average action magnitude $m^{(i)}$. We denote the normalized actions by $\tilde{a}^{(i)}_{t+j}=a^{(i)}_{t+j}/m^{(i)}$. The sampled action chunks can be viewed along two complementary axes. The temporal axis of each action chunk captures C1, while the sample axis across repeated action chunks captures C2. Along the temporal axis, we measure the mean squared changes between consecutive actions within each sampled chunk, defined as
\begin{equation}
T_{\text{temp}}(s_t) = \frac{1}{k} \sum_{i=1}^k \frac{1}{H-1}\sum_{j=1}^{H-1}\|\tilde{a}^{(i)}_{t+j} - \tilde{a}^{(i)}_{t+j-1}\|^2.
\end{equation}
Motivated by AdaFlow's variance-adaptive computation \cite{hu2024adaflowimitationlearningvarianceadaptive}, we measure cross-sample action variance:
\begin{equation}
T_{\text{cross}}(s_t) = \frac{1}{k}\sum_{i=1}^{k}\frac{1}{H}\sum_{j=0}^{H-1}
\big\| \tilde{a}_{t+j}^{(i)} - \bar{\tilde{a}}_{t+j} \big\|^2,
\end{equation} where $\bar{\tilde{a}}_{t+j} = \frac{1}{k}\sum_{i=1}^{k} \tilde{a}_{t+j}^{(i)}$ is the mean action at step $t+j$ over $k$ samples. In practice, we use $k=2$ as two samples are sufficient to reflect the cross-sample divergence.

Before deployment in a new environment, we perform a lightweight calibration by rolling out the base policy and computing $T_{\text{temp}}$ and $T_{\text{cross}}$ at every timestep for a small number of episodes to collect a reference set. The collected statistics form two empirical reference distributions for the environment. During execution, the trigger values $T_{\text{temp}}(s_t)$ and $T_{\text{cross}}(s_t)$ are converted into their corresponding percentiles, $\tau_{\text{temp}}(s_t)$ and $\tau_{\text{cross}}(s_t)$, within their respective reference distributions. The overall trigger score is then defined as
\begin{equation}
\tau(s_t) = \max\bigl(\tau_{\mathrm{temp}}(s_t),\; \tau_{\mathrm{cross}}(s_t)\bigr).
\end{equation}
The trigger follows an OR rule, activating when either percentile exceeds the threshold:
$\tau(s_t) > \tau_0,$
where $\tau_0$ is the only trigger hyperparameter that decides how strict the trigger is. Since the trigger correlates well with the critical states, calibrating $\tau_0$ is straightforward and a few rollout episodes are sufficient.

\subsection{Latent World Model Verifier}

When the Two-Axis Trigger is activated, we perform additional sampling to obtain the remaining $K-k$ action chunk candidates, resulting in a total of $K$ candidates $\{a_{t:t+H}^{(i)}\}_{i=1}^K$. Unlike search-based planning with latent world models \cite{zhou2025dinowmworldmodelspretrained, assran2025vjepa2selfsupervisedvideo}, our verifier ranks action chunk candidates by predicting delta progress without the need for a goal image. This makes it compatible with real-time VLA inference.

\paragraph{Architecture.}

We instantiate the verifier using the Joint-Embedding Predictive Architecture (JEPA) world model framework \cite{assran2025vjepa2selfsupervisedvideo,murlabadia2026vjepa21unlockingdense}, which consists of a pretrained visual encoder and an action-conditioned autoregressive predictor. The encoder compresses the current observation into a latent representation, while the predictor models how the latent state evolves conditioned on robot actions.

To adapt the pretrained world model to our setting, we introduce embodiment-specific state encoders and action encoders together with sub-chunk rollouts. For each embodiment, the encoders project the raw robot states and actions into the predictor embedding dimension. Each action chunk with horizon $H$ is uniformly divided into $M$ sub-chunks, enabling multi-step latent rollouts. This design provides denser supervision than predicting only the state at the end of the full action chunk, while reducing inference overhead and mitigating error accumulation compared with predicting a state after every individual action.

Building upon the latent world model, we introduce a delta progress prediction branch. Unlike latent state prediction, which models task-agnostic environment dynamics, delta progress prediction estimates how much the executed action sub-chunk advances the specified task. Consequently, it requires task-specific information beyond the current observation, robot state, and action. For example, the same state transition corresponds to positive progress for the task ``closing the closet'', but negative progress for ``opening the closet''. We therefore introduce a task-specific learnable query token to condition delta progress prediction.

Following the autoregressive paradigm, the predictor first imagines the next latent state $\hat{z}$ and then predicts the corresponding progress increment $\Delta\hat{p}$ for each input action sub-chunk. An attention mask prevents latent state prediction from attending to the task query. Otherwise, task-specific information could leak into the latent dynamics model, encouraging the predictor to exploit task semantics instead of learning task-agnostic environment dynamics. This design cleanly separates task-conditioned progress estimation from task-agnostic latent dynamics modeling, allowing the pretrained world model to be adapted for verification while preserving its original modeling objective.

\begin{figure*}[t!]
    \centering
    \includegraphics[width=1\linewidth]{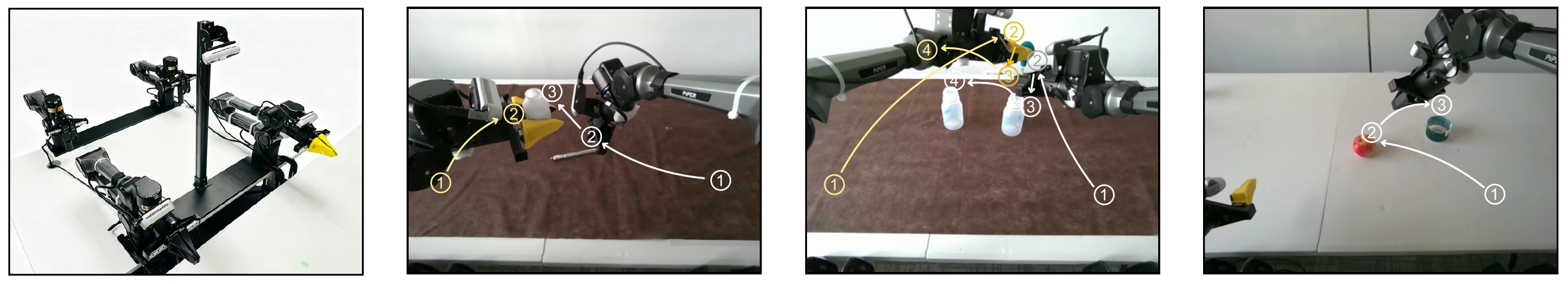}

    \caption{\textbf{Real-world experimental setup.} We conduct the real-world experiments on an AgileX PiperX dual-arm robot, and design three fine-grained tasks, namely, pen insertion, pipetting and apple placement.}
    \label{fig:real_world}

\end{figure*}

\paragraph{Training.}
The verifier is jointly optimized for latent prediction and progress estimation, with progress estimation as the primary objective. Across the \(M\) action sub-chunks, the progress loss combines a Huber regression loss \cite{huber1964robust} with a pairwise ranking loss:
\begin{equation}
\begin{aligned}
&\mathcal{L}_{\text{prog}}
= \sum_{i=1}^{M} \Bigg[
\frac{1}{|\mathcal{B}_i|}
\sum_{k\in\mathcal{B}_i}
L_{\delta}\bigl(\Delta \hat{p}^{(k)}_i,
\Delta p^{(k)}_i\bigr)
\\&+\frac{\lambda_{\text{rank}}}{|\mathcal{P}_i|}
\sum_{(k,l)\in\mathcal{P}_i}
\max\Bigl(
0,\,
\begin{aligned}
&-\bigl(\Delta \hat{p}^{(k)}_i-\Delta \hat{p}^{(l)}_i \bigr)
\\&+\alpha\bigl(\Delta p^{(k)}_i- \Delta p^{(l)}_i\bigr)
\end{aligned}
\Bigr)
\Bigg],
\end{aligned}
\end{equation}
where \(\Delta\hat{p}^{(k)}_i\) and \(\Delta p^{(k)}_i\) denote the predicted and target progress for sample \(k\) at sub-chunk \(i\), respectively. Here, \(\mathcal{B}_i\) is the batch; \(\mathcal{P}_i\) contains same-task pairs with \(\Delta p_i^{(k)}>\Delta p_i^{(l)}\), following DreamAvoid \cite{fan2026dreamavoidcriticalphasetesttimedreaming}. Empty pair sets contribute zero ranking loss. The coefficient \(\lambda_{\text{rank}}\) balances regression accuracy and ranking consistency, while \(\alpha\) controls the target-difference-dependent ranking margin.
 Huber regression fits progress targets, while ranking preserves within-task ordering. The absolute progress is labeled autonomously with Robometer \cite{liang2026robometerscalinggeneralpurposerobotic}, and then the delta progress is calculated based on the result.
The visual prediction loss is defined following V-JEPA 2-AC \cite{assran2025vjepa2selfsupervisedvideo},
\begin{equation}
\mathcal{L}_{\text{vis}} =
    \sum_{i=1}^{M}
    \|\hat{z}_{t+i \cdot \frac{H}{M}} - z_{t+i \cdot \frac{H}{M}} \|_{1}.
\end{equation}
The total training loss is computed as
\begin{equation}
    \mathcal{L} = \mathcal{L}_{\text{prog}} + \lambda_{\text{vis}}\mathcal{L}_{\text{vis}}.
\end{equation}

We use both teacher forcing and self-forcing \cite{huang2026self} during the training stage to balance learning efficiency and inference distribution alignment.

\paragraph{Inference.}

At inference time, the predictor conditions exclusively on its previously predicted latent frames and autoregressively rolls out over the $M$ sub-chunks of each candidate action chunk. The verifier thus produces $M$ delta progress estimates for each candidate. The robot then executes the action chunk with the largest cumulative predicted progress, computed as the sum of these $M$ estimates.

To keep verification practical for real-world deployment, three latency optimizations are applied cumulatively: (O1) reusing the KV cache of the verifier throughout the autoregressive rollout, (O2) optimizing the inference pipeline by overlapping independent computations, including the parallel generation of candidate action chunk samples and performing latent feature encoding in parallel with the action generation of the VLA policy, and (O3) PyTorch-level optimization, including pre-compilation.
\section{Experiments}
\label{sec:experiments}

We organize our experiments around the core question: \emph{how} the verifier should be designed.
We first evaluate the mechanism in three settings: (1) real-world fine-grained manipulation tasks, validating the complete system under practical latency and hardware constraints, (2) RoboTwin 2.0 clean, studying the trade-off between inference efficiency and performance, and (3) RoboTwin 2.0 clean2random, showcasing the generalization ability of the verifier under randomized settings. We conduct ablation studies to assess the model design.
We additionally evaluate on LIBERO across two base policies and 40 tasks. Section~\ref{sec:trigger-analysis} analyzes the trigger; Appendix~\ref{exp:trigger_exp} provides evaluation details.

\begin{table*}[t!]
\centering
\small
\setlength{\tabcolsep}{3pt}

\caption{\textbf{Comparison of different repeated-sampling-and-selection methods across tasks.}}
\label{tab:real_results_wide}
\begin{tabular}{lcccccc}
\toprule
\multirow{2}{*}{Method} & \multicolumn{2}{c}{Validation Metrics} & \multicolumn{4}{c}{\makecell{Completion Score\\(\%) $\uparrow$}} \\
\cmidrule(lr){2-3} \cmidrule(lr){4-7}
 & Rank Loss $\downarrow$ & Huber Loss $\downarrow$ & Pen Insertion & Pipetting & Apple Placement & Average \\
\midrule
$\pi_{0.5}$ \cite{intelligence2025pi05visionlanguageactionmodelopenworld}& -- & -- & 45.0 & 52.5 & 55.0 & 50.8 \\
+ RoboMonkey \cite{kwok2025robomonkeyscalingtesttimesampling}& 1.19 & 0.37 & 49.0 & 55.5  & 63.5 & 56.0 \\
+ GPC-RANK \cite{qi2026inferencetimeenhancementgenerativerobot}& 1.18 & 0.30 & 46.0 & 54.5 & 58.0 & 52.8 \\
\rowcolor{gray!15}
+ \textbf{Ours} & \textbf{1.09} & \textbf{0.25} & 60.0 & 63.5 & 73.0 & \textbf{65.5} \\
\bottomrule
\end{tabular}

\end{table*}

\subsection{Real-world Experiments}
\label{exp:real_world}
\paragraph{Setup.}
We conduct real-world experiments on the AgileX PiperX robotic embodiment, equipped with one front camera (RealSense D455) and two wrist cameras (RealSense D435), as shown in Figure~\ref{fig:real_world}. To evaluate methods in different situations, we design three tasks with varying arm configurations (one single-arm and two dual-arm) and key challenges.
The tasks are pen insertion (pick up the pen and insert it into the yogurt bottle), apple placement (pick up the apple and place it into the container), and pipetting (transfer liquid from the right bottle to the left with a pipette), shown in \Cref{fig:real_world}. The detailed task configurations are provided in Appendix~\ref{appendix:real_world}.

\paragraph{Implementation Details.} For each task, we collect 50 demonstration episodes to fine-tune the base policy. We then collect another 50 policy rollout episodes per task, label them using Robometer \cite{liang2026robometerscalinggeneralpurposerobotic}, and train one verifier for all tasks in the environment on a single H800 GPU. The same protocol is used in the simulation experiments. Each method is evaluated over 50 trials per task in the real-world, RoboTwin 2.0, and LIBERO experiments. The test-time deployment is based on one RTX 4090 GPU, with the policy model and the verifier running on the same device. During testing, we set the trigger threshold to a certain value, where verification is turned on during contact-rich phases. We use $k=2$ samples for Two-Axis Trigger and $K=10$ samples for the verifier.

\paragraph{Metrics.} We utilize three metrics: two validation losses and one closed-loop completion score. Specifically, \textit{rank loss} measures the model's ability to rank the action chunk samples within one task. \textit{Huber loss} measures how well the model regresses delta progress. \textit{Completion score} reflects the system's closed-loop performance on the real-world tasks, where each stage is assigned a predefined score; details are provided in Appendix~\ref{appendix:real_world}.

\paragraph{Baselines.} We compare against the following methods: (1)~Base policy ($\pi_{0.5}$) with single sampling;
(2)~VLM-based verifier following RoboMonkey \cite{kwok2025robomonkeyscalingtesttimesampling}, with the training objective adapted to delta progress estimation;
(3)~Pixel-level world model verifier following GPC-RANK \cite{qi2026inferencetimeenhancementgenerativerobot}.

\begin{figure}[t!]
    \centering
    \includegraphics[width=1\linewidth]{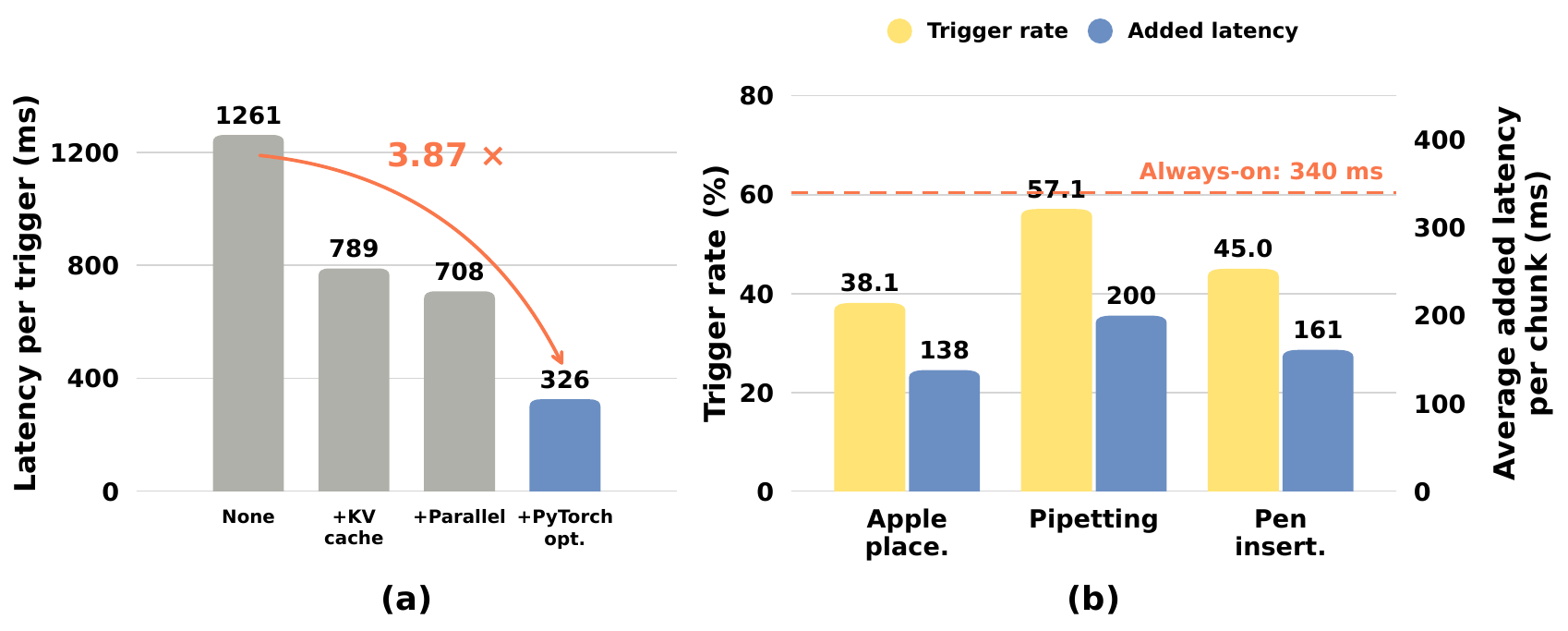}

    \caption{\textbf{Latency analysis.} \textbf{(a) Additional verification latency per triggered decision for 10 samples under successive optimizations.} The three optimization stages cumulatively reduce latency. KV caching provides the largest latency reduction.
    \textbf{(b) Trigger rate (left axis) and the resulting average added latency per chunk (right axis) on the three real-world tasks.}}
    \label{fig:latency}

\end{figure}

\paragraph{Results.} The results are shown in Table~\ref{tab:real_results_wide}. Our method improves the real-world completion score by 14.7 percentage points, outperforming all baselines. Our three-level optimization reduces the latency to 326 ms per triggered decision (trigger latency excluded), shown in \Cref{fig:latency}. RoboMonkey performs competitively on tasks requiring only coarse-scale localization (\textit{e.g.}, apple placement), but struggles with fine-grained manipulation, likely because its VLM lacks an explicit understanding of world dynamics. GPC-RANK achieves strong performance when estimating progress from real observations; however, its performance degrades on test-time rollout frames, indicating that the test-time generation quality becomes a bottleneck.

\subsection{Simulation Results}
To provide reproducible results, we evaluate the combination of the trigger and the verifier in simulation environments, including RoboTwin 2.0 \cite{chen2025robotwin20scalabledata} and LIBERO \cite{liu2023liberobenchmarkingknowledgetransfer}.

\begin{figure}[t]

    \centering
    \includegraphics[width=1\linewidth]{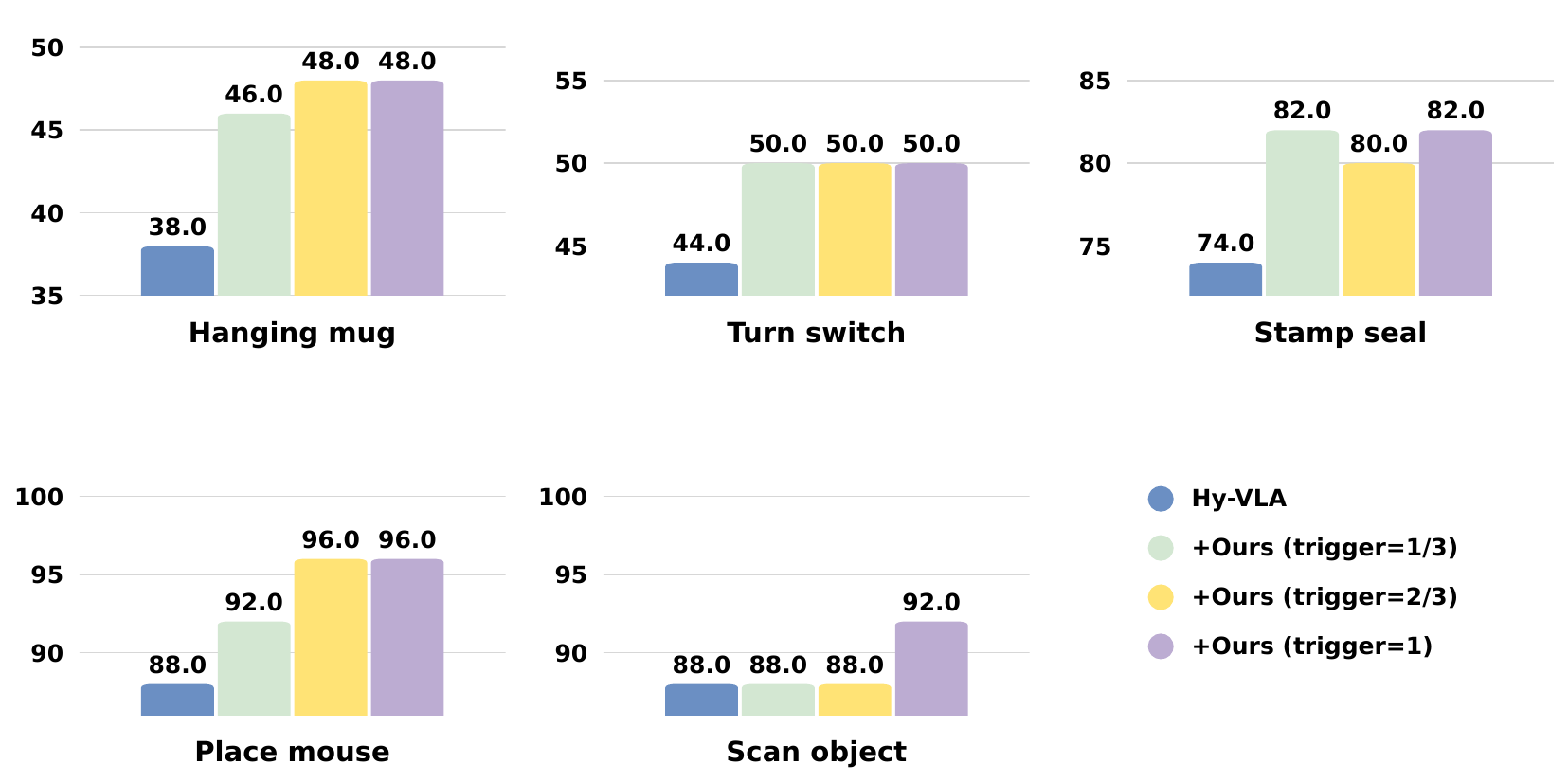}

    \caption{\textbf{Results on RoboTwin 2.0 Clean with Varying Trigger Thresholds.} For $\text{trigger}=r$, we adjust a general trigger threshold for all the tasks, so that approximately an $r$ fraction of decision steps trigger verification.}
    \label{fig:robotwin}

\end{figure}

\paragraph{RoboTwin 2.0.} RoboTwin 2.0 is a benchmark for bimanual manipulation with both clean and randomized settings. For RoboTwin 2.0, we use Hy-Embodied-0.5-VLA \cite{zhang2026hyembodied05vlavisionlanguageactionmodelsrealworld} as the base policy. We evaluate our method on five fine-grained tasks that have relatively low baseline success rates. In the clean setting, we sweep a trigger threshold for all the tasks so that different fractions of decision steps trigger verification. Results show that (1) always-on verification improves the success rate on all five tasks; (2) the trigger could reduce computation while retaining much of the success-rate improvement, as shown in the first four tasks in Figure~\ref{fig:robotwin}. In the Clean2Random setting, both the policy and verifier are trained exclusively on clean data and evaluated in randomized environments with always-on verification. \Cref{tab:robotwin_random} shows that our verifier can boost the base policy in randomized settings by 6.8 percentage points, showcasing generalization capability.

\begin{table}
\centering
\small
\setlength{\tabcolsep}{2pt}
\caption{\textbf{Results on RoboTwin 2.0 Clean and Clean2Random with always-on verification.} Success rates are reported in percent.}
\label{tab:robotwin_random}
\begin{tabular}{lcccccc}
\toprule
Method & Mug & Switch & Seal & Mouse & Scan & Avg. \\
\midrule
\multicolumn{7}{l}{\textit{RoboTwin 2.0 Clean}} \\
\ \ Hy-VLA & 38.0 & 44.0 & 74.0 & 88.0 & 88.0 & 66.4 \\
\ \ + RoboMonkey & 34.0 & 48.0 & 70.0 & 88.0 & 86.0 & 65.2 \\
\ \ + GPC-RANK & 42.0 & \textbf{50.0} & 78.0 & 88.0 & 86.0 & 68.8 \\
\rowcolor{gray!15}
\ \ + \textbf{Ours} & \textbf{48.0} & \textbf{50.0} & \textbf{82.0} & \textbf{96.0} & \textbf{92.0} & \textbf{73.6} \\
\midrule
\multicolumn{7}{l}{\textit{RoboTwin 2.0 Clean2Random}} \\
\ \ Hy-VLA & 16.0 & 50.0 & 72.0 & 80.0 & \textbf{96.0} & 62.8 \\
\ \ + RoboMonkey & 22.0 & 44.0 & 78.0 & 86.0 & 94.0 & 64.8 \\
\ \ + GPC-RANK & 22.0 & 50.0 & \textbf{80.0}& 82.0 & 90.0 & 64.8 \\
\rowcolor{gray!15}
\ \ + \textbf{Ours} & \textbf{26.0} & \textbf{56.0} & 78.0 & \textbf{92.0} & \textbf{96.0} & \textbf{69.6} \\
\bottomrule
\end{tabular}

\end{table}

\paragraph{LIBERO.}
\label{sec:libero}

We evaluate our method on LIBERO \cite{liu2023liberobenchmarkingknowledgetransfer}, using $\pi_{0.5}$ and StarVLA-GR00T \cite{community2026starvlalegolikecodebasevisionlanguageaction} as the base policies. As shown in Table~\ref{tab:LIBERO}, the results indicate that our method improves performance on LIBERO-10 suite, while maintaining strong performance on other suites. Each policy uses its own verifier weights, shared across all 40 tasks but not across the two policies.

\begin{table}[htbp]
\small
\centering
\setlength{\tabcolsep}{3pt}
\caption{\textbf{LIBERO results.} Our method boosts the base policy, especially on LIBERO-10, the task suite that has not been saturated.}
\label{tab:LIBERO}
\begin{tabular}{lcccc c}
\toprule
Method & Spatial & Object & Goal & 10 & Avg. \\
\midrule
$\pi_{0.5}$ & \textbf{98.8} & 98.2 & 98.0 & 92.4 & 96.9\\
~~ + RoboMonkey & 98.6 & \textbf{99.0} & 97.6  & \textbf{94.0} & 97.3 \\
~~ + GPC-RANK & 97.6 & 96.8 & 97.8  & 89.0 & 95.3 \\
~~ + Ours & \textbf{98.8} & 98.8 & \textbf{98.6}  & \textbf{94.0} & \textbf{97.6} \\
\midrule
StarVLA-GR00T & 98.0 & \textbf{99.4} & 96.4 & 91.0 & 96.2 \\
~~ + RoboMonkey & 98.2 & \textbf{99.4} & 96.0  & 91.6 & 96.3 \\
~~ + GPC-RANK & \textbf{98.6} & 99.2 & 96.6  & 92.4 & 96.7 \\
~~+ Ours & 98.4 & 99.0 & \textbf{97.8} & \textbf{93.6} & \textbf{97.2}\\
\bottomrule
\end{tabular}

\end{table}

\subsection{Ablation Study}
In this section, we discuss the design of the verifier, and specifically answer the following question: Do \textit{latent world modeling} and \textit{pretrained latent dynamics} improve progress estimation?
We conduct two sets of ablation studies. First, we assess the role of latent world modeling by removing the visual prediction objective. Second, we investigate whether pretraining the predictor on the general-purpose DROID dataset \cite{khazatsky2024droid} leads to better performance.

\subsubsection{Effect of World Modeling.}

Does the test-time improvement stem from latent world modeling, or merely from a strong visual encoder? To answer the question, we implement a variant that predicts only delta progress, without jointly predicting future visual latents. Therefore, the variant degenerates into an autoregressive transformer without world modeling.
We train the variant on the same three real-world tasks. As shown in \Cref{tab:ablation}, the completion score drops from 65.5\% to 55.2\%. These results suggest that without the dense world modeling objective, the model struggles to learn informative dynamics, leading to substantially weaker performance.

\subsubsection{Effect of Predictor Pretraining}
Our verifier builds on a frozen pretrained V-JEPA 2 encoder, which provides a rich visual prior learned from web-scale video data. In addition, the action-conditioned predictor is pretrained on 23k trajectories from the DROID dataset, offering a strong dynamics prior for robotic manipulation. To assess whether this predictor pretraining is beneficial for knowledge transfer, we train a variant without pretraining where the predictor is initialized randomly.
In Table~\ref{tab:ablation}, the variant shows lower completion scores on real-world tasks, especially on the tasks with stronger generalization requirements, \textit{e.g.} apple placement. In contrast, on the pipetting task, where bottle locations remain almost fixed and generalization is less critical, the performance degradation is relatively minor.

\begin{table}[h]
\centering
\small
\setlength{\tabcolsep}{1.25pt}
\caption{\textbf{Ablation on world modeling and predictor pretraining.} Completion scores (\%) are reported overall and per task. The visual encoder remains pretrained in all variants.}
\label{tab:ablation}
\begin{tabular}{lcccccc}
\toprule
Method & Rank $\downarrow$ & Huber $\downarrow$ & Pen $\uparrow$ & Pipetting $\uparrow$ & Apple $\uparrow$ & Avg. $\uparrow$ \\
\midrule
\rowcolor{gray!15}
VeriVLA & \textbf{1.09} & \textbf{0.25} & \textbf{60.0} & \textbf{63.5} & \textbf{73.0} & \textbf{65.5} \\
\ \ w/o WM & 1.21 & 0.26 & 52.0 & 55.5 & 58.0 & 55.2 \\
\ \  w/o pretrain & 1.17 & \textbf{0.25} & 54.0 & 61.0 & 63.5 & 59.5 \\
\bottomrule
\end{tabular}
\end{table}

\subsection{Trigger Analysis}
\label{sec:trigger-analysis}

We compare the Two-Axis Trigger with a trained DINOv2--MLP baseline, action chunk entropy (ACE) \cite{liang2026adaptiveactionchunkinginferencetime}, and its temporal and cross-sample components. Our trigger requires no training. Baseline configurations and evaluation details are provided in Appendix~\ref{exp:trigger_exp}.

\paragraph{Metrics.} \textit{Status AUC} measures alignment between trigger scores and contact-based reference labels from privileged simulator states. \textit{Variance Correlation} measures Spearman's $\rho$ between trigger scores and reference action variance estimated from $K$ policy samples per state. These diagnostics assess contact relevance and candidate diversity; full definitions appear in Appendix~\ref{exp:trigger_metric}.

\begin{table}[t]

    \centering
    \setlength{\tabcolsep}{3pt}
    \small

    \caption{\textbf{Trigger performance analysis.} $k$ denotes the number of action samples; latency is additional per decision.}
\label{tab:trigger}
\begin{tabular}{lccc}
        \toprule
        Method
    & \multicolumn{1}{l}{\makecell{Status \\ AUC $\uparrow$}}
    & \multicolumn{1}{l}{\makecell{Variance \\ Correlation $\uparrow$}}
    & \multicolumn{1}{l}{\makecell{$k$ / add.\\latency (ms) $\downarrow$}} \\
        \midrule
        random & 0.487 & -- & -- \\
        DINO & 0.702 & 0.630 & \textbf{0 / 0} \\
        ACE & 0.588 & 0.716 & 2 / 14 \\
        $T_{\text{temp}}$ & 0.704 & 0.789 & \textbf{1 / 0} \\
        $T_{\text{cross}}$ & 0.694 & 0.877 & 2 / 14 \\
        Ours & \textbf{0.727} & \textbf{0.887} & 2 / 14 \\
        \bottomrule
    \end{tabular}

\end{table}

\paragraph{Results.} In \Cref{tab:trigger}, the combined trigger achieves a Status AUC of 0.727 and a Variance Correlation of 0.887, exceeding both individual components. The temporal component has higher Status AUC (0.704 versus 0.694), while the cross-sample component has higher Variance Correlation (0.877 versus 0.789), suggesting complementary signals. Our trigger uses two samples with 14 ms of additional latency per decision.

Variance Correlation thus evaluates whether a $k=2$ trigger tracks the $K$-sample reference. \Cref{fig:robotwin} separately evaluates the closed-loop success--computation trade-off.

\section{Limitations and Future Work}

A key limitation of repeated sampling and selection is the difficulty of steering a pretrained VLA model. Although the verifier can rank sampled action candidates, its effectiveness depends on the diversity and quality of the proposals generated by the underlying policy. The diversity of repeated sampling with different random seeds is still relatively limited.
Another limitation is the use of a frozen visual encoder. While freezing the encoder reduces training cost and preserves pretrained representations, these representations are not explicitly optimized for manipulation progress estimation. They may overlook subtle changes in object configurations, contact conditions, or task completion that are essential for distinguishing promising actions from ineffective ones.
Future work can investigate learning an encoder--predictor pair for progress estimation within a JEPA framework by considering progress estimation as an optimization goal.

\section{Conclusion}

We have introduced an adaptive test-time verification framework for exploiting the stochasticity of flow-based VLA policies. Our training-free two-axis trigger identifies critical states from within-chunk motion variation and cross-sample disagreement, invoking additional sampling only when it is likely to be useful. A JEPA-based latent world model then ranks candidate action chunks by jointly predicting action-conditioned future latents and delta task progress, avoiding pixel reconstruction and goal-image dependence. Experiments in both real-world and simulation demonstrate consistent gains across tasks, environments, and base policies. On three real-world tasks, our method increases the average completion score from 50.8\% to 65.5\% while adding 326 ms for sampling and verification per triggered decision, excluding the 14 ms trigger overhead. Ablation studies further confirm the benefits of latent dynamics modeling and pretraining. Taken together, these results establish adaptive latent verification as an effective and practical approach to improving frozen generative VLAs at deployment.

\clearpage
\bibliographystyle{verivla_cvpr}
\bibliography{aaai2027}

\clearpage
\appendix
\onecolumn
\section*{Appendix}

\section{Extended Related Work}
\label{appendix:related_work}
In this section, we provide an extended review of prior work.

\subsection{Test-time Verification}
Existing methods differ in how the verifier is constructed. \emph{Model-based verifiers} train a separate scoring model: RoboMonkey \citep{kwok2025robomonkeyscalingtesttimesampling} fine-tunes a 7B VLM on synthetic action preferences, CoVer \citep{kwok2026scaling} argues that scaling such verifiers beats scaling policy learning, and V-GPS \citep{nakamoto2024steering} re-ranks samples with an offline-RL value function. \emph{Model-free} methods instead read the signal off the policy itself, via masked-condition divergence \citep{jang2026verifierfree} or proximity to the discrete action tokens \citep{zhao2026actiondraftverifyselfverifying}; both measure agreement rather than progress, and neither reasons about the physical consequences of a candidate. \emph{World-model verifiers} close this gap by imagining outcomes, but do so in two stages: GPC-RANK \citep{qi2026inferencetimeenhancementgenerativerobot} and DreamAvoid \citep{fan2026dreamavoidcriticalphasetesttimedreaming} synthesize future pixels before scoring them, while FOREWARN \citep{wu2025foresight} rolls out latents and then aligns a per-task finetuned VLM to narrate them. VeriVLA keeps the world model but removes the second stage, reading delta progress directly from predicted latents. Combined with a two-axis action-based trigger, verification is invoked adaptively.

\begin{table*}[htbp]
\centering
\small

\setlength{\tabcolsep}{3pt}
\caption{\textbf{Comparison of test-time verification methods.}}
\label{tab:comparison}
\begin{tabular}{lcccc}
\toprule
\textbf{Method} & \textbf{Trigger} & \makecell{\textbf{World} \\ \textbf{Modeling}} & \makecell{\textbf{Verification} \\ \textbf{Pipeline}} & \makecell{\textbf{Prediction}\\ \textbf{Target}} \\
\midrule
RoboMonkey  & always-on & $\times$ & one-stage & action preference \\
FOREWARN  & always-on & \checkmark & two-stage & latent, VLM narration \\

GPC-RANK  & always-on & \checkmark & two-stage & pixel, reward \\
ADV  & always-on & $\times$ & one-stage & proximity to discrete action tokens \\
MG-Select  & always-on & $\times$ & one-stage & masked-condition divergence \\

DreamAvoid  & vision-based & \checkmark & two-stage & pixel, delta progress \\
VeriVLA & action-based & \checkmark & one-stage & latent, delta progress \\
\bottomrule
\end{tabular}

\end{table*}

\subsection{Other Test-time Enhancement}
In addition to test-time verification, we provide an overview of other existing test-time enhancement methods, including test-time RL, adaptive action chunking, and online monitoring.

\textbf{Test-time training and RL.} Test-time training methods, such as EVOLVE-VLA \citep{bai2025evolvevlatesttimetrainingenvironment} and TT-VLA \citep{liu2026ontheflyvlaadaptationtesttime}, perform online policy optimization during deployment using dense reward signals. Rather than relying on external reward estimation, T$^2$VLA \citep{chen2026trustinstinctsconfidencedriventesttime} derives intrinsic rewards from the policy's own confidence to enable self-supervised adaptation. These methods update policy parameters and therefore require reward access, gradient computation, and tolerance for transient regression during adaptation.

\textbf{Adaptive action chunking.} A parallel line of work asks not \emph{which} action to take but \emph{how} to emit it. BID \citep{liu2025bidirectional} resamples action chunks under backward coherence and forward contrast to trade off consistency against reactivity. RTC \citep{black2025realtime} targets the latency gap by generating the next chunk while the current one executes, freezing the actions guaranteed to run and inpainting the remainder. AAC \citep{liang2026adaptiveactionchunkinginferencetime} adaptively sets the chunk size from action entropy, and DVAC \citep{feng2026denoising} triggers replanning from denoising variance. The underlying intuition traces back to AdaFlow \citep{hu2024adaflowimitationlearningvarianceadaptive}, which links the conditional variance of the flow-matching loss to ODE discretization error and reduces to a one-step generator when the action distribution is unimodal. Our trigger shares this variance-based intuition, but these methods reshape the execution or decoding schedule of a single trajectory and never assess whether the emitted actions make task progress; our trigger instead gates a verifier that can reject and replace a chunk rather than merely smoothing or resizing it.

\textbf{Online monitoring and failure detection.} Runtime monitors aim to predict failure before it becomes irrecoverable. Sentinel \citep{agia2024unpacking} splits failures into erratic modes, detected by statistical temporal consistency of action-chunk distributions (STAC), and task-progression modes, detected via VLM video QA; the two detectors together catch 18\% more failures than either alone. SAFE \citep{gu2026safe} shows that VLA internal features carry task-generic success/failure information and trains a lightweight scalar failure predictor on them, using conformal prediction for a calibrated time-varying threshold; VLA-FAIL \citep{seligmann2026vla} pairs a last-layer Mahalanobis distance with an action-chunk consistency term to remove the need for failure data. Closest to our setting, Foresight \citep{zhang2026foresight} extracts action-conditioned latents from V-JEPA\,2-AC and predicts a per-timestep failure score, showing that world-model features generalize across policies better than policy-internal ones.

\section{Training Details}
\subsection{Hyperparameters}
We list the training hyperparameters in \Cref{tab:hyperparams}.

\begin{table}[htbp]
\centering

\caption{\textbf{Training hyperparameters and resource configuration.} The table specifically shows the real-world adaptation experiments. The actual training epochs and duration can largely depend on the scale of the dataset.}
\label{tab:hyperparams}
\begin{tabular}{lc}
\hline
\textbf{Hyperparameter} & \textbf{Value} \\
\hline
Learning rate & $5 \times 10^{-4}$ \\
Weight decay & $0.04$ \\
Batch size & 32 \\

Optimizer & AdamW \\
JEPA loss weight & $1$ \\
Rank loss weight & 1 \\
Crop size & $256 \times 256$ \\
Patch size & 16  \\
Tubelet size & 2\\
Number of sub-chunks & 5 \\

\hline
\textbf{Resource} & \textbf{Specification} \\
\hline
GPU model & 1 NVIDIA H800 (80 GB) \\
\hline
\end{tabular}

\end{table}

\section{Trigger Experiments}
\label{exp:trigger_exp}

\subsection{Evaluation Metrics}
\label{exp:trigger_metric}
Our trigger aims to detect critical task states or states where the policy is uncertain. Since there is no ground truth for when to trigger, we propose two metrics in simulation environments to evaluate how well the trigger value aligns with our setting: (1) \textit{Status AUC} is the AUC-ROC between the trigger value and a rule-based \textit{critical label}. The label is a binary indicator obtained by thresholding temporal proximity to \textit{critical moments}, which are read from the simulator and defined as the moments when task-related objects switch between contact and non-contact, with the gripper or another task-related object. This measures how well the trigger identifies contact-relevant stages that we hypothesize to be most challenging. It serves as a fair metric, since the criterion is computed solely from simulator states rather than the visual states or policy actions that are used by triggers, and it avoids manual annotation. The critical-label construction is detailed in Section~\ref{appendix:status_AUC}.
(2) \textit{Variance Correlation} measures Spearman's $\rho$ between trigger scores and a $K$-sample reference variance. At each evaluated state, we draw $K$ action chunk samples from the policy and compute their variance as an empirical ground-truth reference for policy variability. This $K$-sample Monte Carlo reference differs from the $k=2$ sample variance used by the trigger at deployment. The metric evaluates whether a trigger computed with few samples tracks variability estimated from the larger sample set; it does not correlate the trigger with its own two-sample variance.

\subsection{Critical-label Construction}
\label{appendix:status_AUC}

To evaluate whether a trigger signal at a state $V_\text{trigger}(s_t)$ identifies moments
that genuinely require more compute, we construct a reference label
$\text{crit}(t) \in [0, 1]$ that is independent of the action or vision. We find the \textit{critical moments}
from privileged simulator state (object/gripper pose, joint angles, contact) using a rule: we mark frames where gripper-object or object-object contact state flips (grasp / release) for any task-relevant object. For each support moment, we assign a Gaussian kernel function. The \textit{critical index} for each frame is then computed as the sum of all overlapping kernel contributions, clipped to a maximum of 1.

After labeling an episode with the continuous critical index, we compute the trigger value for each chunk $a_{t:t+H}$ and take the maximum critical index within that window. Thresholding the latter at 0.5 yields the binary label, and the trigger value serves as the score; we then compute the ROC curve and the Status AUC. We visualize two tasks from LIBERO and RoboTwin for reference:
\begin{figure*}[t]
    \centering
    \includegraphics[width=\textwidth]{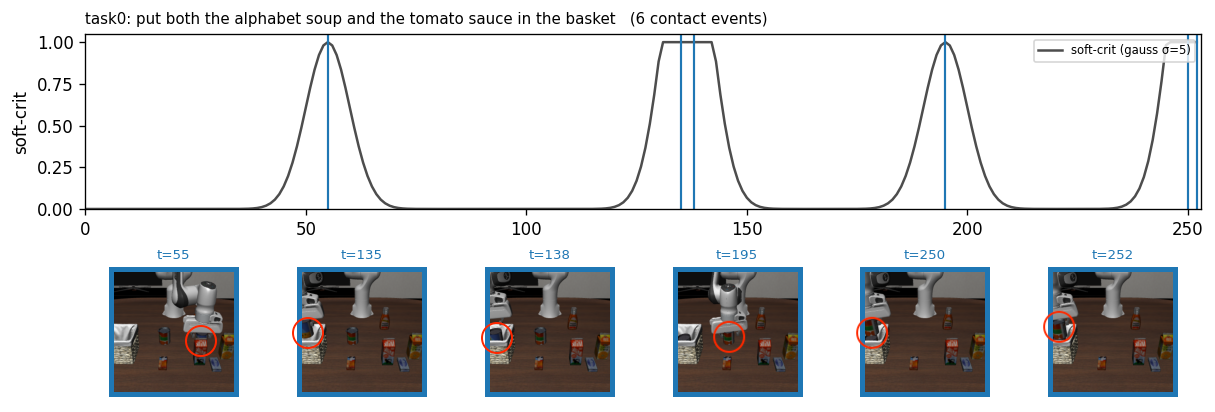}

\caption{\textbf{Critical Index and Critical Moments in LIBERO.}}
\label{fig:critical-phase-example}
\end{figure*}
\begin{figure*}[t]
    \centering
\includegraphics[width=0.8\textwidth]{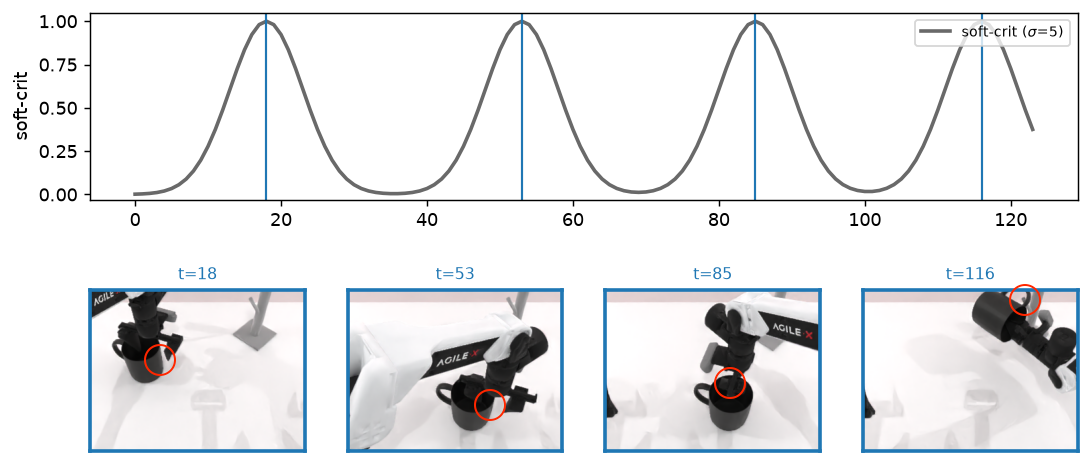}

\caption{\textbf{Critical Index and Critical Moments in RoboTwin.}}
\label{fig:critical-phase-robotwin}
\end{figure*}

\subsection{Baselines} We compare the proposed Two-Axis Trigger against representative prior methods as well as its individual components: (1)~A lightweight MLP trained on frozen DINOv2 features, as a representative method for a vision-based baseline \cite{fan2026dreamavoidcriticalphasetesttimedreaming}. We train the model with the signals read from the simulator, and evaluate it together with the other methods on the perturbation set from LIBERO-Plus \cite{fei2025liberoplusindepthrobustnessanalysis}.
(2)~Action chunk entropy (ACE) \cite{liang2026adaptiveactionchunkinginferencetime}. Although originally designed for adaptive action chunking rather than the trigger, it provides an action-based baseline.
(3)~The temporal trigger $T_\text{temp}$ and cross-sample variance trigger $T_\text{cross}$, components of our Two-Axis Trigger.

The quantitative comparison is reported in Section~\ref{sec:trigger-analysis} and \Cref{tab:trigger}.

\section{Real-world Setting}
\label{appendix:real_world}
\subsection{Tasks}
We conduct experiments on three real-world tasks, pen insertion, pipetting, and apple placement. \Cref{tab:stage-scores} defines the completion score of each task.
\begin{table*}[t]
\centering

\setlength{\tabcolsep}{3pt}
\caption{\textbf{Completion score breakdown for the three real-world tasks.}}
\label{tab:stage-scores}
\begin{tabular}{lcp{0.65\textwidth}c}
\toprule
Task & Stage & Completion criterion & Score \\
\midrule
\multirow{2}{*}{Pen Insertion}
 & 1 & Grasp the pen & 0.5 \\
 & 2 & Insert the pen into the yogurt bottle & 0.5 \\
\midrule
\multirow{4}{*}{Pipetting}
 & 1 & Move the right hand to the correct position at the right bottle & 0.25 \\
 & 2 & Successfully aspirate water with the left hand & 0.25 \\
 & 3 & Move the right hand to the correct position at the left bottle & 0.25 \\
 & 4 & Successfully dispense water with the left hand & 0.25 \\
\midrule
\multirow{2}{*}{Apple Placement}
 & 1 & Grasp the apple & 0.5 \\
 & 2 & Place the apple into the container & 0.5 \\
\bottomrule
\end{tabular}
\par\smallskip
\begin{minipage}{0.98\textwidth}
\footnotesize
For apple placement, if successful placement requires multiple attempts, the placement stage receives 0.25 points instead of 0.5 points.
\end{minipage}

\end{table*}

\subsection{Computational Cost}
We report the inference computational cost in \Cref{tab:latency-gpu}, including latency and GPU usage.

\begin{table}[t]
\centering

\setlength{\tabcolsep}{3pt}
\caption{\textbf{Latency and GPU memory usage breakdown.}}
\label{tab:latency-gpu}
\begin{tabular}{p{0.65\columnwidth}r}
\toprule
\multicolumn{2}{c}{\textbf{Latency}} \\
Component & \makecell{Latency\\(ms)} \\
\midrule
VLA policy & 82 \\
Two-Axis Trigger ($k{=}2$) & 14 \\
Additional sampling ($K{-}k{=}8$ more candidates) & 80 \\
Verifier forward (all $K{=}10$ candidates) & 246 \\
\midrule
Total latency (deactivated) & 96 \\
Total latency (activated) & 422 \\
Additional latency (verification only) & 326 \\
\midrule
\multicolumn{2}{c}{\textbf{GPU Memory Usage}} \\
Component & \makecell{GPU Memory\\(GB)} \\
\midrule
VLA policy & 9.8 \\
Verifier forward (all $K{=}10$ candidates) & 8.7 \\
\midrule
Combined GPU memory & 18.5 \\
\bottomrule
\end{tabular}

\end{table}
\end{document}